# A NOVEL DESIGN SPECIFICATION DISTANCE (DSD) BASED K-MEAN CLUSTERING PERFORMACE EVALUATION ON ENGINEERING MATERIALS' DATABASE

Doreswamy,
Department of Computer Science,
Mangalore University, Mangalagangotri-574 199
doreswamyh@yahoo.com

Hemanth. K. S
Department of Computer Science,
Mangalore University, Mangalagangotri-574 199
reachhemanmca@gmail.com

**ABSTRACT:**
Organizing data into semantically more meaningful is one of the fundamental modes of understanding and learning. Cluster analysis is a formal study of methods for understanding and algorithm for learning. K-mean clustering algorithm is one of the most fundamental and simple clustering algorithms. When there is no prior knowledge about the distribution of data sets, K-mean is the first choice for clustering with an initial number of clusters. In this paper a novel distance metric called Design Specification (DS) distance measure function is integrated with K-mean clustering algorithm to improve cluster accuracy. The K-means algorithm with proposed distance measure maximizes the cluster accuracy to 99.98% at P = 1.525, which is determined through the iterative procedure. The performance of Design Specification (DS) distance measure function with K - mean algorithm is compared with the performances of other standard distance functions such as Euclidian, squared Euclidean, City Block, and Chebshew similarity measures deployed with K-mean algorithm. The proposed method is evaluated on the engineering materials database. The experiments on cluster analysis and the outlier profiling show that these is an excellent improvement in the performance of the proposed method.

## 1. INTRODUCTION

Advancement in sensing and digital storage technologies and their dramatic growth in the applications ranging from market analysis to scientific data explorations have created many high-volume and high dimensional data sets. Most of the data stored in electronic media have influenced the development of efficient mechanisms for information retrieval and automatic data mining tools for effective classification and clustering of high-dimensional data. In addition to this, the exponential growth of high-dimensional data requires advanced Data Mining (DM) methodology to automatically understand process and summarize data. DM is a process of extracting previously unknown, potentially useful and ultimately understandable knowledge from the high volume of data. Data mining techniques can be broadly classified into two major categories (**Jiawei Han et al, 2012**): (i) explorative or descriptive task, characterizes the general properties or structures of the high dimensional data without any pre-defined models or hypothesis, and (ii) Predictive or confirmatory or inferential task, confirms the validity of hypothesis/model or a set of inferences given the availability of data. Many statistical approaches have been proposed to explore data analysis, such as variance analysis, linear/multilinear regression, discriminant analysis, correlation analysis, multidimensional scaling, factor analysis, Principal Component Analysis (PCA) and for Cluster analysis (**Eduardo R. Hruschka et al, 2009; Manish Verma et al, 2012**).

Data mining concerned with predictive data analysis involves different levels of learning, such as (i) supervised (CLASSIFICATION) learning, involves with only labeled data (training patterns) and predicts the behavior of the unseen data sets and (ii) unsupervised (CLUSTERING) learning, involves with only unlabeled data, and (iii) semi-supervised learning some time also called as hybrid setting, involves partial labeled and unlabeled data sets for understanding the hidden behavior of the data sets. Clustering is a more difficult and challenging problem than classification.

### 1.1. Data clustering
Data clustering, also called cluster analysis, is the discovery of semantically meaningful grouping of natural data sets or patterns or objects. It can also be defined as a given representation of a set of objects that are divided into k groups based on similarity measure so that the similarity between any two objects within a group, $k_i$ is maximized and the similarity between any two objects within any two groups' $k_i$ and $k_j$ is minimized. Cluster analysis is prevalent in any discipline that involves analysis of multivariate data. It has been used for under laying structures to gain insight into data, generate hypothesis, detect anomalies and identify silent features (**31**), for a natural classification to identify the degree of similarity among engineering materials(**12,19**), and for compression to organize the data and summarizing it through cluster prototypes ( ).

### 1.2. Historical development of Cluster
The development of clustering methodology is a truly interdisciplinary endeavor, taxonomist, social scientist, psychologists, biologists, mathematicians, engineers, computer scientist, medical researchers and others who collect and process real data, have all contributed to clustering methodology. According to JSTOR (2009), data clustering is first appeared in the title of a 1954 article dealing with anthropological data. Data clustering is also known as Q-analysis, clumping, and taxonomy depending on the field where it is used (**Jain A K, 2009**). Data clustering algorithms have been extensively studied in data mining (Han **and Kamber, 2012**).

The Cluster can be broadly classified into two categories, hierarchical and partitional clustering. Hierarchical clustering is recursively found nested clusters either in agglomerative mode or in the divisive mode (Two-down) mode, while partitional algorithms find the clusters simultaneously a partition of data and do not impose a hierarchical structure. The most well-known algorithms are single –link and complete line. The most popular and the simplest partition algorithm is K-mean. K-mean is rich and has diverse history of the past five decades (**Jain A K, 2009**). Even through K-mean was first was discovered ove the 50 years ago, it is still one of the most widely used algorithms for clustering because of its simplicity, effeiciency and empeical success.

The rest of the paper is organized as follows. Section 2 describes a brief review of K-mean Clustering Algorithm. A K - mean algorithm with different standard distance functions are described in the section 3. A novel design specification distance measure is described in section 4. Proposed experimental results of DSD with k-mean clustering are discussed in section 5. Comparison of the proposed method with K-mean with other distance measure functions on engineering materials data sets is done in the section 6. Conclusion and future scopes are given section 7.

## 2. A BRIEF REVIEW OF K-MEAN CLUSTERING ALGORITHM

Now a day K-mean algorithm is most widely used algorithm in data mining applications (**T. Velmurugan and T. Santhanam, 2011; Manish Verma et al, 2012**). It is a simple, scalable, easily understandable and can be adopted to deal with high dimensional data. Let $X = \{x_i, i = 1, 2...N\}$ be the n-dimensional points to be clustered into a set of K-Clusters, $C = \{ C_i, i= 1, 2..k\}$. The K - Mean algorithm finds partitions such that the squared error between the empirical mean of a cluster $C_i$ and the points, $x_i$, in the cluster $C_i$ is minimized. Let $\mu_k$ be the mean of the cluster $C_k$, the squared error between $\mu_k$ and the points, $x_i$, in $C_k$ is defined as

$$E(C_k) = \sum_{x_i \in C_k} \|x_i - \mu_k\|^2 \quad (1)$$

The goal of K-mean is to minimize the sum of the squared error over all K clusters.

$$min_{\mu_1...\mu_k} E(C) = \sum_{k=1}^{n} \sum_{x_i \in C_k} \|x_i - \mu_k\|^2 \quad (2)$$

Where $\|x_i - \mu_k\|^2$ is the Euclidean distance similarity measure function.

**Algorithm 1:** K-means

**Inputs:** $k \geq 1$
1. Select initial cluster prototypes $\mu_1, \mu_2, \ldots, \mu_k$
2. **Repeat**
3. **for all** $x_i \in X$ **do**
4. **for all** $\mu_i$ **do**
5. Compute the similarity
$$E(C_k) = \sum_{x_i \in C_k} \|x_i - \mu_k\|^2$$
6. **end for**
7. Assign an object $x_i$ to cluster $C_i$ of which
$$\min E(c_k) = \sum_{x_i \in c_k} \|x_i - \mu_k\|^2$$
8. **end for**
9. **for all** $\mu_i$ **do**
10. Update $\mu_i$ as the centroid of cluster Ci
11. **end for**
12. **Until** converges is attained to zero.

The Euclidean distance metric is the commonly used distance measure with K-mean algorithm and is not efficient for clustering high dimensional data sets as it bypasses some data sets, which are relatively considered as outliers by other few reputed and frequently used distance measures with K-mean algorithm for clustering. Some time, the outliers considered by the Euclidean distance function may also be relevant data sets for decision making. The inclusion of such patterns for maximizing cluster accuracy is a task of the proposed method.

## 3. K-MEAN CLUSTER WITH SIMILARITY FUNCTIONS

Distance measure involved a superlative task in clustering data objects. Several reputed and frequently used distance measures shown in Table 1 are proposed for data clustering in different applications. Each measure has its own advantages and drawbacks (**Eduardo et al., 2009**) that depend on the type of data sets being used. It is well-known that some measures are more suitable for gene clustering in Bioinformatics (**D T Pham et al., 2007**), and for text clustering and document categorization (**Todsanai et al., 2009**). Therefore, depending upon the problem and database, the distance measuring functions contribute a major role in the cluster algorithm (**Guadalupe et al.2008**). Anil Kumar Patidar, et al (2012) used four standard similarity measure functions such as Euclidean, Cosine, Jaccard and Person correlation function in SNN clustering algorithm on a synthetic dataset, KDD Cup'99, Mushroom data set and some randomly generated database. In SNN technique generally data must be cleaned in order to find desired cluster. Here, they are inserting un-clustered data to desired core cluster discovered by SNN algorithm. Ultimately, they suggested in their studies that Euclidean measure performed well in SNN algorithm comparable to other three measures. Kuang-Chiung Chang, et al(2005), studied and compared the data clustering performances of four similarity measures city block(L1-norm), Euclidean(L2-norm), normalized correlation coefficient and simplified grey relational grade on QRS complexes and found that a simplified grey relational grade distance measure shown better performance on medical data sets. AnkitaVimal et al (2008), a brief study of various distance measures and their effect on different clustering algorithms is carried out in this article. With the help of k-mean matrix partitioning and dominance based clustering algorithms; Euclidean distance measure and other four distance measure were studied to analyze their performance by accuracy of various techniques using synthetic datasets. Real-world data sets of cricket and synthetic datasets from Syndeca software were used for cluster analysis. In this study it is found that the Euclidean distance measure performs better than the other measures. Doreswamy et al, (2010) carried out a study on different similarity measure functions and proposed a new one called exponential similarity measure function for the selection of engineering materials based on the input design requirements.

Distance measure functions frequently used with k-means algorithm and tabulated in Table1 are standard distance metric functions.

Table 1: Represents the few frequently used distance measures for clustering

| Distance Measure | Formula | Description |
|---|---|---|
| Euclidean distance | $D_{Ed}(X,Y) = \sqrt{\sum_{i=1}^{n}(x_i - y_i)^2}$ | Where $x_i$ and $y_i$ represent the N dimension and $D_{Ed}$ is a symmetric matrix. The most commonly used metric special cases of the Minkowski distance at n=2 tend to from hyper-spherical clusters. |
| City Block Distance | $D_{cBd}(X,Y) = \sum_{i=1}^{n}|x_i - y_i|$ | Special case of Minkowski distance at n=1 tend to form hyper-rectangular clusters. |
| Chebysher Distance | $D_{Cd}(X,Y) = max|x_i - y_i|$ | Also known as the L1-distance, between two points in a Euclidean space with a fixed Cartesian coordinate system defined as the sum of the lengths of the projections of the line segment between the points onto the coordinate axes |
| Squared Euclidean Distance | $D_{cBd}(X,Y) = \sum_{i=1}^{n}|x_i - y_i|^2$ | Squared Euclidean Distance is not a metric as it does not satisfy the triangle inequality; however, it is frequently used in optimization problems in which distances only have to be compared. |
| Minkowski Distance | $D_{Md}(X,Y) = \left[\sum_{i=1}^{n}|x_i - y_i|^p\right]^{\frac{1}{p}}$ | Where p is the user parameter; Its typically used p values are 1 and 2. |

Development of novel distance measure functions that maximize the cluster accuracy of k-mean algorithm is an open task for research in different applications of data mining **[10]**. This motivates the development of a new distance measure, DSD measure, with K-mean to maximize the data clustering accuracy on engineering materials data sets.

## 4. DESIGN SPECIFICATION DISTANCE MEASURE

From scientific and mathematical point of view, distance is defined as a quantitative degree that enumerates the logical separation of two objects represented by a set of measurable attributes/characteristics. Measuring a distance between two data points is a core requirement for several data mining tasks that involve distance computation. A novel distance function is proposed to measure the logical representation of two engineering materials based on their design specification/ Characteristics. Therefore, this measure is called as Design specification distance measure. Well distance is an amount that reflects the strength of the relationship/similarity between two data items. Without losing of generality a distance only needs to operate on finite datasets. Formally, distance is a function D with no-negative real values, defined on the cartesian product X×X of a set R.

Where R is the real number. For all x, y, z in R, this function is required to satisfy the following condition:
D (x, y) = 0, If only if x=y. (The identity axiom)
D(x, y) ≥ 0, (non-negativity, or separation axiom)
D (x,y)=D(y,x). (The symmetry axiom)
D (x,y) + D(y,z) ≥D(x,z). (The triangle inequality)

The number of points in a data set is denoted by N. Each point is denoted by $x_i$ and $y_i$ so on. D denotes the set of dimensions, and D $(x_i, y_i)$ represent the subsets of dimensions of the points $x_i$ and $y_i$ respectively.

Design specification distance measure function is defined on two data points $X = (x_1, x_2, ..., x_n)$ and $Y = (y_1, y_2, ..., y_n) \in R^n$, is defined as:

$$D(X,Y) = \left[\sum_{i=1}^{n}|x_i - y_i|^2\right]^{\frac{p}{3}} \quad (3)$$

$$= \begin{cases} \text{if p=3; } D(X,Y) \text{ is a square Euclidean distance.} \\ \text{if p=1.5; } D(X,Y) \text{ is a Euclidean distance function} \\ \text{if p=1.523; } D(X,Y) \text{ is DSD distance} \end{cases}$$

Where i=1....n, p is a user defined parameter.

The p is a positive integer constant, $1 \leq p \leq 3$. This function satisfies the condition of the non-negativity, the identity of indiscernible, symmetry conditions and triangle inequality of distance measuring definition. If p = 3 the distance is squared Euclidian distance, and the function is Euclidean distance for **p = 1.5.** Using the proposed Design Specification distance measure in k-mean algorithm, a novel data clustering model is designed and implemented for the grouping engineering materials data sets. And K-mean algorithm with a distance measure functions in table 1 are compared with the proposed Design Specification (DS) distance metric on engineering materials data sets.

## 5. EXPERIMENTAL RESULTS

This section discusses the experimental results of the proposed method on the engineering materials database and compares its clustering accuracy along with outlier profiling with other methods. In order to evaluate the proposed method, prototype software is designed and developed and implemented in Visual C# .Net and installed on the Computer system having Intel Core Due Processor with 4 GB RAM and 500 GB Hard disk. This software provides Graphical User Interface (GUI) as gateway to provide interfaces between the database and as well as users. Data mining engine implemented for clustering is deployed in between GUI and Database. A typical GUI for accessing materials data set from the object-oriented database model is shown in figure 1.

### 5.1 Engineering Materials Database
Engineering materials are the scientifically designed materials that are falling into two major categories such as Matrix and Reinforcements. Matrix materials are further classified into three categories such as Polymer, Metal and Ceramic. The reinforcement materials are classified into short, medium and long fibers based the shear strength of matrix and critical length of fiber materials. A composite material is combination of matrix and as well as reinforcement material that yields with better mechanical strength compare to the basic engineering materials. Each type has different properties namely Thermal, Chemical, Mechanical, Electrical, Magnetic, Physical, Environmental, Acoustical and Manufacturing process properties. These properties/attributes are modeled with Object-Oriented features and called as Object-Oriented Data Model for Engineering materials data sets. The attribute values of each data set is sampled from the materials handbook, research papers and the globally available materials data sets at www.matweb.com, which is currently being used for materials selection in engineering composite materials in manufacturing industries. Typical Object-Oriented Data Model of Advance Engineering materials data set (Doreswamy et al, 2012) is implemented for mining aspects.

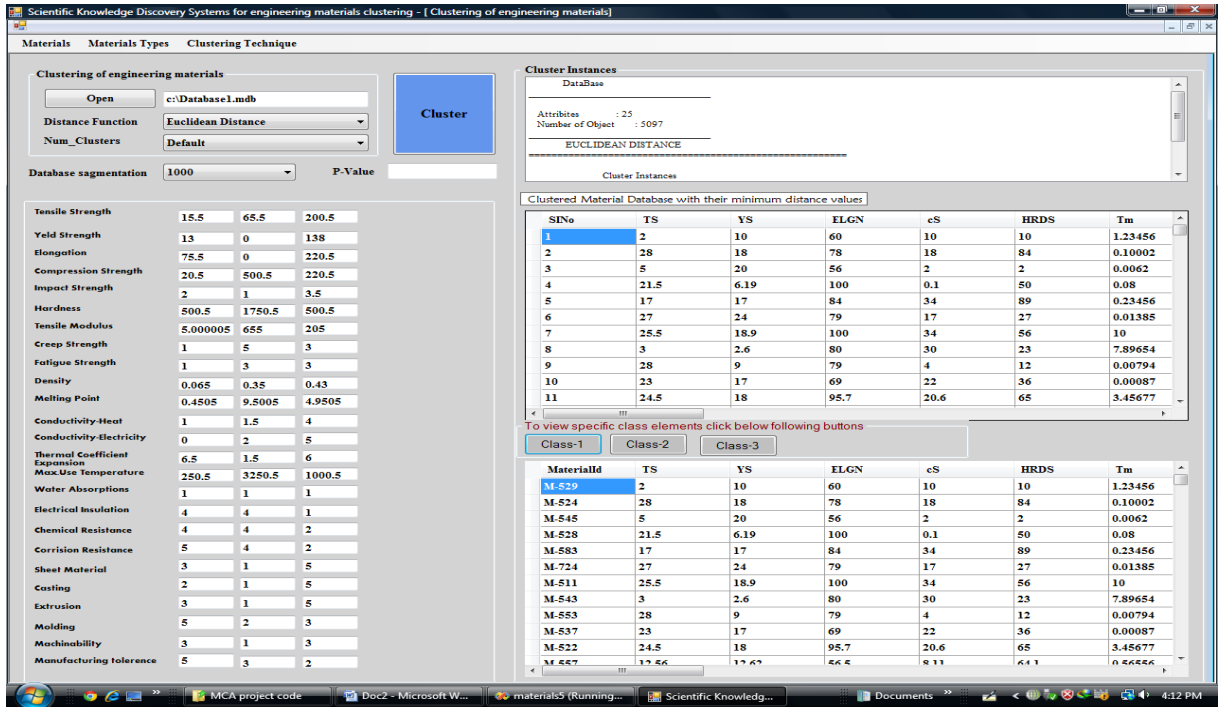

Figure 1: GUI for accessing materials data sets from the Object-Oriented Data Model

### 5.2 Data Normalization

Engineering materials data organized and stored in the Object-Oriented data model are not uniformly distributed and spread over different ranges of attributes. If the Tensile Modules of a Metal material say A has values from $4*10^4$ to $9*10^7$ GMp and the Tensile modules of another metal say B has values from $4*10^4$ to $9*10^5$ GMp, then An influence on the distance function will usually be overpowered by A's influence. K-mean algorithm with distance function results with poor clustering accuracy dominated by outlier if the data sets are not standardized to a uniform range. Possible outlier data sets are avoided by employing Min-Max normalization **(Han and Kamber, 2012)** to transform a value v of a numeric attribute A to $v^|$ in the range [0, 1] by computing.

$$V^| = \frac{v - min_A}{max_A - min_A} \quad (4)$$

where $min_A$ and $max_A$ are the minimum and maximum values of attribute A.

### 5.3. Evaluation of Novel Distance measure

The proposed method is evaluated on Object-oriented engineering materials database containing 5097 data sets. The attribute values of each sampled data set in the database are lie in the standard range of attributes ( ) of different classes, where each class refers to a type of engineering materials namely polymer, ceramic and metal. The K - mean algorithm requires an initial number of clusters, K, to cluster the engineering materials database. In this experiment, K value is always set to 3 as the database contains only three categories of materials in the matrix class. Maximum 25 attribute values are sampled in each K class at each time. The experiment was conducted at 5 different instances with varying data sets { 1000, 2000, 3000, 4000, 5097} for each value of P = {1.0, 1.2, 1.34, 1,42, 1.45, 1.5, 1.523 ,1.55, 1.56, 3.0}. In each iteration with different instances of data sets, the data sets were clustered into three major classes namely Class-1(Polymer), Class-2(Ceramic) and Class- 3(Metal).The same experiment was conducted for varying values of P. The outcomes of the iterative experiments were observed and tabulated in the Table 2. It is found from these observations that the clustering accuracy and the outlier percentage are varying as the number of data set considered increasing in each instance for clustering. The clustering accuracy and the outlier percentage of the proposed algorithm are computed by

$$Cluster\ Accurcy\ \% = \frac{Number\ of\ clustered\ Data}{Total\ number\ of\ Dataset} \times 100 \quad (5)$$

And

$$Outlier = \frac{(Total\ number\ of\ Dataset - Number\ of\ Clustered\ Data)}{Total\ number\ of\ Dataset} \times 100 \quad (6)$$

**Table 2: Data clustering observations of different of data instances with varying P values**

| Varying P Values | Data sets clustered (K = 3) | Data sets considered at different instances | | | | |
|---|---|---|---|---|---|---|
| | | 1 | 2 | 3 | 4 | 5 |
| | | 1000 | 2000 | 3000 | 4000 | 5097 |
| P=1 | Cluster-1 | 314 | 653 | 905 | 1202 | 1502 |
| | Cluster-2 | 329 | 615 | 886 | 1201 | 1511 |
| | Cluster-3 | 281 | 619 | 919 | 1182 | 1584 |
| | Cluster Accuracy % / Outliers % | 92.4% / 7.6% | 94.35% / 5.65% | 90.33% / 9.66% | 89.62% / 10.3% | 91.94% / 8.04% |
| P=1.2 | Cluster-1 | 320 | 663 | 933 | 1254 | 1590 |
| | Cluster-2 | 311 | 611 | 921 | 1235 | 1564 |
| | Cluster-3 | 287 | 622 | 958 | 1233 | 1611 |
| | Cluster Accuracy % / Outliers % | 91.8% /8.2% | 94.8% /5.2% | 93.73% /6.26% | 93.05% /6.95% | 93.48% /6.51% |
| P=1.34 | Cluster-1 | 324 | 664 | 937 | 1274 | 1605 |
| | Cluster-2 | 333 | 613 | 924 | 1254 | 1598 |
| | Cluster-3 | 288 | 624 | 962 | 1261 | 1643 |
| | Cluster Accuracy % / Outliers % | 94.5% /5.5% | 95.05% /4.95% | 94.1% /5.9% | 94.72% /5.2% | 95.075% /4.92% |
| P =1.42 | Cluster-1 | 328 | 668 | 953 | 1284 | 1622 |
| | Cluster-2 | 339 | 621 | 981 | 1299 | 1640 |
| | Cluster-3 | 292 | 639 | 984 | 1281 | 1674 |
| | Cluster Accuracy % / Outliers % | 95.9% /54.1% | 96.4% /3.6% | 97.266% /2.73% | 96.6% /3.4% | 96.84% /3.58% |
| P=1.45 | Cluster-1 | 335 | 679 | 974 | 1316 | 1664 |
| | Cluster-2 | 312 | 632 | 975 | 1321 | 1665 |
| | Cluster-3 | 298 | 644 | 994 | 1292 | 1688 |
| | Cluster Accuracy % / Outliers % | 94.5% /5.5% | 97.75% /2.25% | 98.1% /1.9% | 98.22% /1.775% | 98.43% /1.569% |
| <span style="color:red">P= 1.5</span> | <span style="color:red">Cluster-1</span> | <span style="color:red">348</span> | <span style="color:red">697</span> | <span style="color:red">993</span> | <span style="color:red">1341</span> | <span style="color:red">1689</span> |
| | <span style="color:red">Cluster-2</span> | <span style="color:red">347</span> | <span style="color:red">649</span> | <span style="color:red">995</span> | <span style="color:red">1344</span> | <span style="color:red">1694</span> |
| | <span style="color:red">Cluster-3</span> | <span style="color:red">300</span> | <span style="color:red">646</span> | <span style="color:red">996</span> | <span style="color:red">1294</span> | <span style="color:red">1690</span> |
| | Cluster Accuracy % / Outliers % | 99.5% /0.5% | 99.6% /0.4% | 99.46% /0.53% | 99.47% /0.525% | 99.52% /0.47% |
| P=1.523 | <span style="color:red">Cluster-1</span> | <span style="color:red">348</span> | <span style="color:red">697</span> | <span style="color:red">986</span> | <span style="color:red">1344</span> | <span style="color:red">1694</span> |
| | <span style="color:red">Cluster-2</span> | <span style="color:red">350</span> | <span style="color:red">652</span> | <span style="color:red">1002</span> | <span style="color:red">1352</span> | <span style="color:red">1704</span> |
| | <span style="color:red">Cluster-3</span> | <span style="color:red">301</span> | <span style="color:red">650</span> | <span style="color:red">1001</span> | <span style="color:red">1302</span> | <span style="color:red">1699</span> |
| | Cluster Accuracy % / Outliers % | 99.9% /0.1% | 99.95% /0.05% | 99.63% /0.366% | 99.95% /0.05% | 99.98% /0.019% |
| P=1.55 | Cluster-1 | 348 | 697 | 995 | 1344 | 1694 |
| | Cluster-2 | 350 | 652 | 1003 | 1353 | 1704 |
| | Cluster-3 | 301 | 650 | 1001 | 1303 | 1701 |
| | Cluster Accuracy % / Outliers % | 99.9% /0.1% | 99.95% /0.05% | 99.96% /0.033% | 100% /00% | 100.039% /0.039% |
| P=1.56 | Cluster-1 | 351 | 701 | 1003 | 1358 | 1714 |
| | Cluster-2 | 350 | 652 | 1003 | 1353 | 1704 |
| | Cluster-3 | 305 | 657 | 1020 | 1327 | 1728 |
| | Cluster Accuracy % / Outliers % | 100.6% /0.6% | 100.5% /0.5% | 100.86% /0.866% | 100.95% /0.95% | 100.43% /0.431% |
| P = 3.0 | Cluster-1 | 348 | 697 | 995 | 1342 | 1691 |
| | Cluster-2 | 349 | 651 | 999 | 1347 | 1698 |
| | Cluster-3 | 308 | 648 | 996 | 1297 | 1693 |
| | Cluster Accuracy % / Outliers % | 100.5% /0.5% | 99.8% /0.2% | 99.66% /0.33% | 99.65% /0.35% | 99.66% /0.33% |

Further it is observed from the Table 2, distribution of data sets to three different classes varying as *p* values get change and subsequently both the cluster accuracy % and the outlier % changes. Behaviors of the discriminated functions in equ.(4) are closely observed and found that when, p = 1.3 and P= 1.5, Design Specification( DS) distance function behaves exactly as Squared Euclidean distance and Euclidean distance functions respectively. However, there is an inclusion of a few common data sets in more than one cluster that results with Cluster Accuracy % greater than 100% and subsequently increases the higher outlier percentage, when **P** takes the values 1.55, 1.56 and 3.0. The accuracy percentage and outer profile of the K-mean algorithm with Design Specification distance function for varying values P on maximum 5097 data sets are shown in the figure 3. The value of P is fixed in 1.523 as it maximizes the cluster accuracy to 99.98% and minimizes the outlier to 0.0196 % for the large data sets considered at a time.

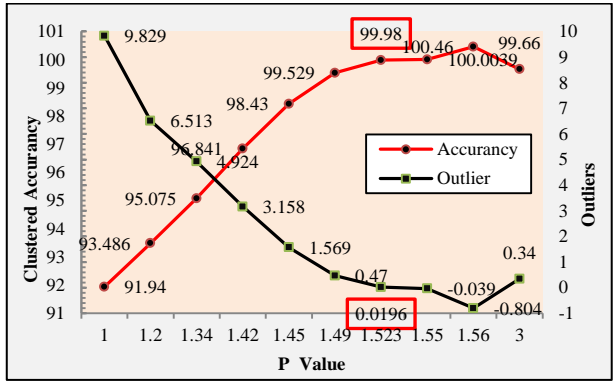

**Figure 3:** Cluster Accuracy % and Outlier % associated with various values of P

As the proposed DS distance function accurately performs well at P = 1.523, the proposed distance function is compared with other distance measure functions with K-mean algorithm.

## 6. EXPERIMENTAL EVALUATION OF DIFFRENT MEASURES

This section compares the proposed distance function, Design Specification (DS) Distance function as discussed in section 4, with other distance functions discussed in this paper. K-mean algorithm was implemented using each of the distance functions: Euclidian, Minkowski, City Block, Chebshew, squared Euclidean, and chebshew distance measures. Each distance function was tested on an engineering materials database. The accuracy and outlier profile of 5 trials of different instances with three classes were reported in the table 3.

**Table 3: Data clustering accuracy and outlier profile of different distance measures**

| Sl.No. | K-mean Algorithm with Different Distance Measure Functions | Data sets clustered (K = 3) | Data sets considered at different instances | | | | |
|---|---|---|---|---|---|---|---|
| | | | 1 | 2 | 3 | 4 | 5 |
| | | | 1000 | 2000 | 3000 | 4000 | 5097 |
| 1 | **Minkowski Distance** | Cluster-1 | 348 | 697 | 992 | 1340 | 1687 |
| | | Cluster-2 | 347 | 648 | 989 | 1318 | 1700 |
| | | Cluster-3 | 299 | 643 | 993 | 1289 | 1684 |
| | | **Cluster Accuracy / Outliers %** | **99.4% /0.6%** | **99.4% /0.6%** | **99.1% /0.86%** | **98.67% /1.325%** | **99.48% /0.51%** |
| 2 | **CityBlock Distance** | Cluster-1 | 346 | 691 | 981 | 1328 | 1653 |
| | | Cluster-2 | 343 | 640 | 976 | 1326 | 1670 |
| | | Cluster-3 | 300 | 647 | 981 | 1276 | 1679 |
| | | **Cluster Accuracy / Outliers %** | **98.9% /1.1%** | **98.9% /1.1%** | **97.93% /2.066%** | **98.25% /1.75%** | **98.138% /1.86%** |
| 3 | **Euclidean Distance** | Cluster-1 | 348 | 697 | 993 | 1341 | 1689 |
| | | Cluster-2 | 347 | 649 | 995 | 1344 | 1694 |
| | | Cluster-3 | 300 | 646 | 996 | 1294 | 1690 |
| | | **Cluster Accuracy / Outliers %** | **99.5% /0.5%** | **99.6% /0.4%** | **99.46% /0.53%** | **99.47% /0.525%** | **99.52% /0.47%** |
| 4 | **Square Euclidean Distance** | Cluster-1 | 348 | 697 | 995 | 1342 | 1691 |
| | | Cluster-2 | 349 | 651 | 999 | 1347 | 1698 |
| | | Cluster-3 | 308 | 648 | 996 | 1297 | 1693 |
| | | **Cluster Accuracy / Outliers %** | **100.5% /0.5%** | **99.8% /0.2%** | **99.66% /0.33%** | **99.65% /0.35%** | **99.66% /0.33%** |
| 5 | **Chebysher Distance** | Cluster-1 | 346 | 692 | 983 | 1326 | 1668 |
| | | Cluster-2 | 346 | 645 | 983 | 1325 | 1665 |
| | | Cluster-3 | 300 | 644 | 984 | 1283 | 1677 |
| | | **Cluster Accuracy / Outliers %** | **99.2% /0.8%** | **99.05% /0.95%** | **98.33% /1.66%** | **98.35% /1.65%** | **98.29% /1.7%** |
| 6 | **Design Specification Distance** | Cluster-1 | 348 | 697 | 986 | 1344 | 1694 |
| | | Cluster-2 | 350 | 652 | 1002 | 1352 | 1704 |
| | | Cluster-3 | 301 | 650 | 1001 | 1302 | 1699 |
| | | **Cluster Accuracy / Outliers %** | **99.9% /0.1%** | **99.95% /0.05%** | **99.63% /0.366%** | **99.95% /0.05%** | **99.98% /0.019%** |

## 6.1. Outlier Profiling

Outlier is a noise or a data set that does not belong any other groups. Maximization of clustering accuracy by minimizing outlier % on any large data set is the objective of any good clustering algorithm. The outlier profiling of each method is done and comparisons are made in the figure 4. The outliers obtained by the proposed method are less compared to other methods.

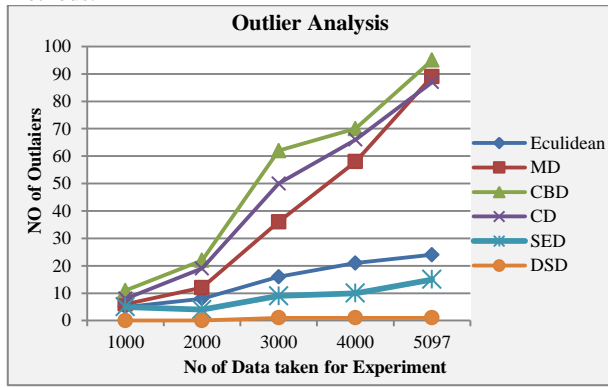

Figure 4: Representing the outlier analysis of data clustered from the six distance measures.

The clustering accuracy % and outlier % of K-mean with different distance measure function are shown in the figure 5. It is found that the proposed Design Specification (DS) distance function on large database containing 5097 datasets, maximizes cluster accuracy to 99.88% by minimizing outlier accuracy to 0.019%.

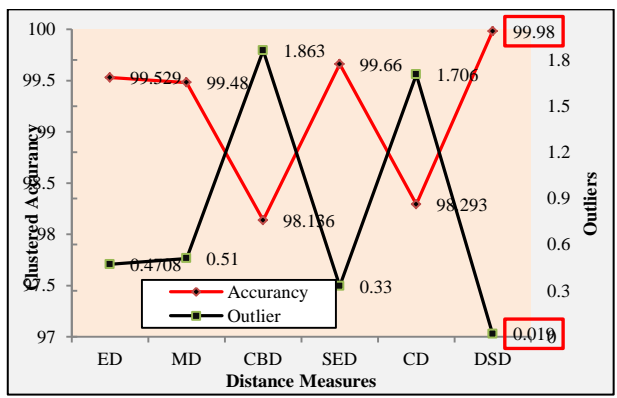

Figure 5: Comparison of Cluster Accuracy and outlier profile of the proposed method with other standard Distance measure functions

## 7. CONCLUSION AND FUTURE SCOPE

From the survey of k-mean clustering algorithm and the experimental results, the following observations were made:

- There is no clustering algorithm that can be universally used to solve all problems. Usually, algorithms are designed with certain assumptions and favor some type of biases.
- With this knowledge, we can say that, there is no best clustering algorithm for all problems even though some comparisons are possible. These comparisons are mostly based on some specific applications, under certain conditions, and the results may become quite different if the conditions change to handle a large volume of data as well as high-dimensional features.
- A distance measuring function is used to measure the similarity among objects, in such a way that more similar objects have lower dissimilarity value. Several distance measures can be employed for clustering tasks. Each measure has its own merit and demerits. The selection of different measures is a problem dependent.
- Here proposed distance measure for k-mean clustering algorithm is in favor of clustering the engineering material's database. Due to variation of the p parameter in the measuring function, at some point say at P = 1.523, it gives the optimal results.
- Hence, choosing an appropriate distance measure for k-mean clustering algorithm can greatly reduce the burden of succeeding designs.

In this paper, a survey of k-mean clustering algorithm in different fields is studied and a novel Design Specification (DS) distance measuring function is proposed with k-mean algorithm for clustering engineering materials database. The proposed algorithm accurately maximizes the cluster accuracy by minimizing outliers. The performance of the proposed method outperformance the other standard methods and can play a critical role in advanced engineering materials design applications.

Further clustering algorithm can be extended to cluster database containing both numeric and ordinal values by employing fuzzy inference rules.